\documentclass[letterpaper, 10 pt, conference]{ieeeconf}  

\IEEEoverridecommandlockouts                              

\makeatletter
\let\NAT@parse\undefined
\makeatother

\usepackage[american]{babel}
\usepackage{mathtools} 
\usepackage{booktabs} 
\usepackage{tikz} 
\usepackage{lineno}

\usepackage{tabularx}
\usepackage{graphicx}
\usepackage{amssymb}
\usepackage{ulem}
\usepackage{multirow}
\usepackage{algorithm}
\usepackage{algorithmic}
\usepackage{siunitx}

\usepackage{xcolor}

\usepackage[parfill]{parskip}
\newcommand{\f}{\zeta}

\newcommand{\ST}[1]{{\color{red}$<$#1$>$}}

\usepackage{hyperref}
\usepackage{cite}

\title{\vspace{0.2cm}\LARGE \bf Acoustic Wave Manipulation Through Sparse Robotic Actuation}

\author{Tristan Shah$^{1*}$, Noam Smilovich$^{1*}$, Feruza Amirkulova$^{2}$, Samer Gerges$^2$, Stas Tiomkin$^{1\dagger}$
\thanks{$^*$ Equal contribution. Listing order is random.}
\thanks{$^{1}$ Computer Science Dept., Whitacre CoE, Texas Tech University; {\tt\{trisshah, noam.smilovich, stas.tiomkin\}@ttu.edu}.}
\thanks{$^{2}$ Mechanical Engineering Dept., Davidson CoE, San Jose State University; {\tt\{samer.gerges, feruza.amirkulova\}@sjsu.edu}.}
\thanks{$^\dagger$ Corresponding Author.}
}

\begin{document}

\maketitle

\thispagestyle{empty}
\pagestyle{empty}

\begin{abstract}
Recent advancements in robotics, control, and machine learning have facilitated progress in the challenging area of object manipulation. These advancements include, among others, the use of deep neural networks to represent dynamics that are partially observed by robot sensors, as well as effective control using sparse control signals. In this work, we explore a more general problem: the manipulation of acoustic waves, which are partially observed by a robot capable of influencing the waves through spatially sparse actuators. This problem holds great potential for the design of new artificial materials, ultrasonic cutting tools, energy harvesting, and other applications. We develop an efficient data-driven method for robot learning that is applicable to either focusing scattered acoustic energy in a designated region or suppressing it, depending on the desired task. The proposed method is better in terms of a solution quality and computational complexity as compared to a state-of-the-art learning based method for manipulation of dynamical systems governed by partial differential equations. Furthermore our proposed method is competitive with a classical semi-analytical method in acoustics research on the demonstrated tasks. We have made the project code publicly available, along with a web page featuring video demonstrations: \url{https://gladisor.github.io/waves/}.

\end{abstract}

\section{Introduction}
\label{sec:intro}

Manipulation has long been a central challenge in robotics \cite{zhu2022_challenges_robotic_manipulation, matas2018_sim_to_real_robotic_manipulation}. Robots are frequently tasked with physically interacting with their environment, be it through picking, placing, or altering objects in a variety of applications, from industrial manufacturing \cite{sanchez2018_industry_robotic_manipulation, mikkel2016_robot_industrial_manufacturing} to healthcare \cite{hamed2012_robot_surgery, boonvisut2013_robot_surgery_sensing}. In this work, we explore robotic manipulation on a class of vastly more complex phenomena: acoustic waves. 

Manipulation of acoustic waves introduces unique challenges on several fronts. Unlike traditional object manipulation, where the robot operates in direct contact with the object, wave manipulation involves indirect influence through intermediary tools such as leveraging the interaction between the scatterers and the incident field \cite{Martin06,amirkulova_2020_the, norris2011_multiple}. Further complicating the task, acoustic waves propagate at an extremely rapid speed \cite{del1972_speed_of_sound} compared to robotic actuation. Lastly, designing an optimal controller for wave manipulation requires knowledge about the state of wave, which is a function of time and space, and usually can be only partially observed. 


Establishing robust control of acoustic waves has wide reaching applications. Robotic systems capable of manipulating waves can lead to groundbreaking technologies, such as 
seismic wave mitigation \cite{chen_2023_artificially,kim_2021_longitudinal} and super-focusing devices for high-resolution ultrasound imaging and surgery \cite{ilovitsh_2018_acoustical,ozcelik_2018_acoustic}. Mastering wave manipulation opens the door to these and many other downstream technologies. 
\begin{figure}[t!]
    \centering
    \includegraphics[width=\linewidth]{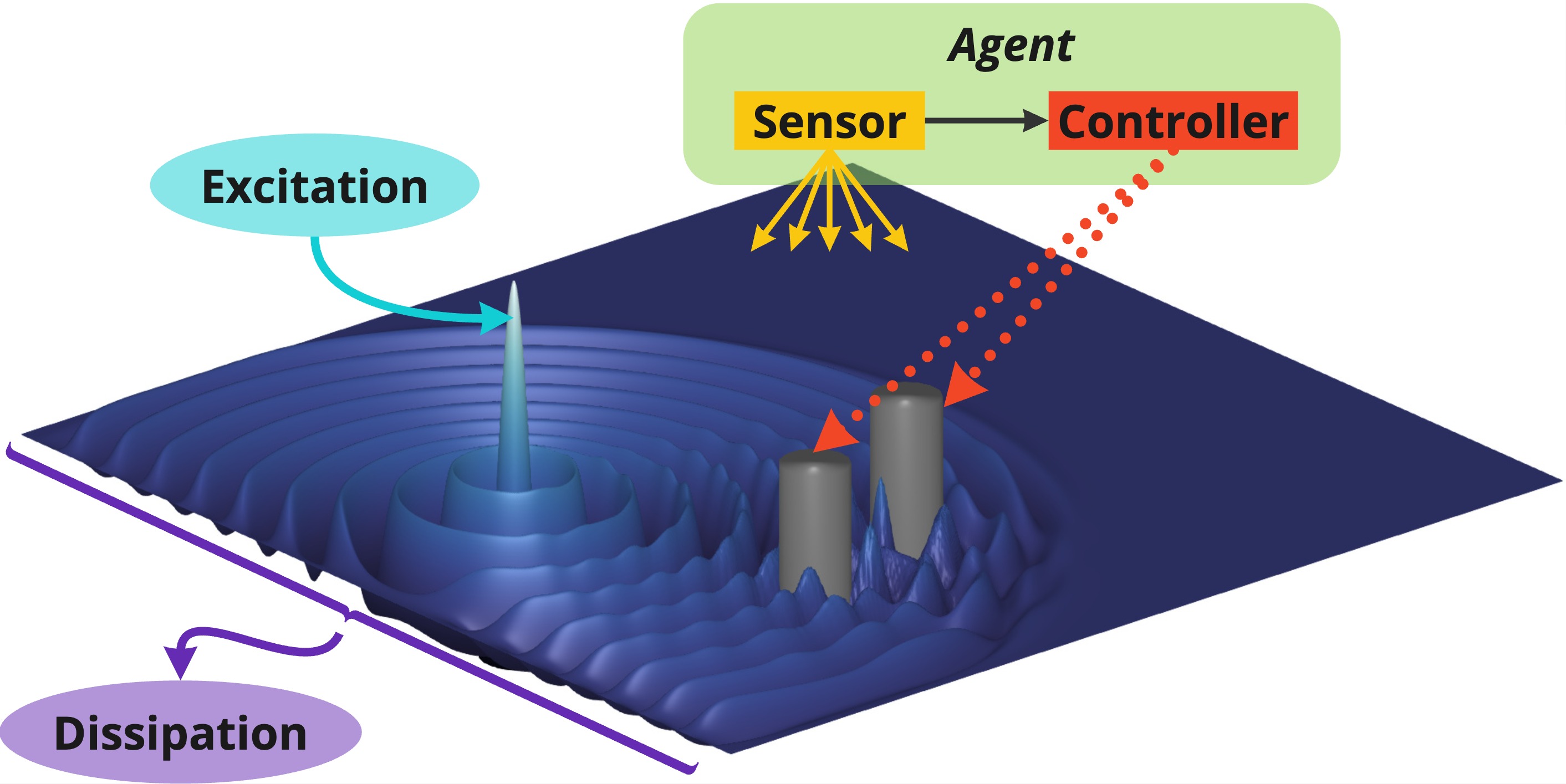}
          \caption{Schematic representation of interaction between an agent and acoustic wave. The agent observes wave through its `Sensor' readings, and affects it back by choosing the optimal locations and radii of the cylindrical scatterers (gray blobs) according to its `Controller' (policy). Wave propagates with the speed of sound according to the Wave PDE, while robot dynamics (actuation of the cylinders) is an order of magnitude slower time scale. `Excitation' represents an exogenous source of acoustic energy (e.g., speaker). The `Dissipation' layer prevents wave reflections from the boundaries and allows for the efficient simulation of open space (infinite) environments. Robot's policy is trained to achieve a desired wave configuration such as focusing energy in a particular location or minimizing total scattered energy.}
    \label{fig:main_scheme}
\end{figure}


Aside from numerical methods like finite difference \cite{kreiss2002difference,leveque2007finite,angel2023efficient} and discontinuous Galerkin \cite{cockburn2012discontinuous,shu2016high}, which rely on computationally intensive simulations, as well as the Reduced Order Model approach \cite{Morris_Amirkizi2023_ROM,Wang2023T_ROM}, typically suited for low-frequency wave components, there are two main approaches for wave manipulation.
The first one assumes an analytical solution to the wave equation, which is a partial different equation (PDE) \cite{courant1967partial}. If such an analytical solution can be obtained, it is used in conjunction with gradient based optimization (GBO) techniques to perform optimization over a system parameters \cite{amirkulova_2020_the,Amirkulova_etal2021, Amirkulova_Gerges_Norris_2022,GergesAmirkulova2024}, towards a desired control objective. However, in general, an analytical solution to a PDE with arbitrary initial and boundary conditions is unknown, which leads to simplifying and limiting assumptions of the problem \cite{amirkulova_2020_the}.





Another approach is based on machine learning and data-driven methods. The existing data-driven methods for PDE control suggest to learn wave dynamics from a large number of samples in order to predict its response to a control signal \cite{werner_2023_learning, bieker_2020_deep}. Subsequently, this model can be used by a predictive controller to select optimal actions. Usually the learning of PDE dynamics is done by way of a black-box neural network model such as a (long short term memory) LSTM model \cite{werner_2023_learning, bieker_2020_deep}, Latent Evolution of PDE (LE-PDE) \cite{wu_2022_learning}, or, more advanced methods such as a Neural Ordinary Differential Equation (NODE) \cite{_2019_neural}. 

The main disadvantage of these black-box models is that there is no guarantee that a model will learn essential properties of the underlying dynamics of a PDE such as forces, energy dissipation, or viscosity. Consequently, due to the uninterpretable nature of black-box models, there is no clear way to represent physical properties of a wave within the model's architecture. That hampers the applicability of black-box models to important real-life applications such as wave propagation in open space, which requires physically meaningful dissipation of  energy.

To address the gaps in the existing methods we design an interpretable framework for realizable control of acoustic waves. Our framework is specifically designed for adaptation to real-world environments, effectively addressing key challenges in wave manipulation. Namely: considering a realizable actuation mechanism which affects the wave in a sparse manner by inducing scattering is employed, actions are selected by an agent on a slower timescale than wave propagation, and observations of the wave state which could be feasibly collected by sensors are provided as input to our model. Furthermore, our framework addresses the lack of interpretability \cite{yu2021_interpretability} of previous-ML based methods by explicitly constraining a learnable latent representation to abide by a one-dimensional (1D) wave equation. By construction, physical meaning is assigned to quantities within the latent space in the form of parameterized initial conditions, boundary conditions, and constraints (explained in Section \ref{sec:propsol}). Moreover, our framework allows for wave control in open spaces, which we achieve by the design of a trainable data-driven dissipation layer. To our best knowledge, the latter has not been demonstrated before, while it is an essential component in acoustics research, known as \textit{perfectly matched layer} (PML) \cite{berenger_1994_a, johnson_2021_notes}.

We demonstrate the application of our framework to two important tasks in acoustics: sound suppression (minimization of scattered energy in the domain) and sound focusing (maximization of sound energy in a desired location). In our experiments we show that our method is superior to a standard ML dynamics model in these tasks, and competitive with a semi-analytical GBO method. Our work is pioneering research in dynamical control of acoustic waves. While this work focuses on establishing control of a particular type of PDE for wave phenomenon, it could be applied to controlling arbitrary systems governed by PDE such as wildfires, pandemics, and eventually climate \cite{miguelngeljavaloyes_2023_a, schneider_1974_climate,majid_2021_analysis}.

\section{Prior Work}
\label{sec:prior}
There have been several attempts to apply machine learning to the manipulation of complex phenomena governed by differential equations. Overall there has been less focus on designing systems that could be realized by real robots. 


One approach used a surrogate dynamics model to enhance reinforcement learning for PDE-governed systems \cite{werner_2023_learning}. The authors did not prioritize realizability in this work, assuming their agent could apply forces across the entire spatial domain of the 1D Kuramoto-Sivashinsky (KS) system—a non-realistic scenario, as fluid manipulation typically involves sparse control through rotating cylinders \cite{bieker_2020_deep} or airfoils \cite{CHEN_XU_LU_2010}. Additionally, their reliance on full state observability, with an LSTM model trained on high-resolution simulation data, limits its applicability to real-world systems, where only sensor data is available.

A realizable approach to using ML for fluid dynamics manipulation was demonstrated through MPC of a Navier-Stokes governed system \cite{bieker_2020_deep}. The authors made specific choices to promote the realizability of their system, such as  using sparse control through rotating cylindrical devices as well as assuming partial observability through sensors. Similar to the example of KS manipulation \cite{werner_2023_learning}, this work also employed an LSTM architecture \cite{bieker_2020_deep}. While the authors demonstrated the success of their method, the sample efficiency of such an architecture may not be optimal for modeling PDE dynamics. Furthermore, the dynamics of an LSTM are not interpretable and do not provide a straightforward pathway to incorporating domain knowledge from fluid dynamics.

Recent works have explored the incorporation of physics knowledge in ML dynamics models \cite{nicodemus_2022_physicsinformed, ericaislanantonelo_2022_physicsinformed}. These works and others \cite{mraissi_2019_physicsinformed} have shown that incorporating such knowledge into the training of neural networks leads to better generalization and sample efficiency. In their work, the authors utilized Physics-Informed Neural Networks (PINNs) \cite{mraissi_2019_physicsinformed} to model the dynamics of simple systems governed by ODE in response to control. They demonstrated that their PINN based dynamics models could effectively control a mechanical multi-link manipulator \cite{nicodemus_2022_physicsinformed}, the Van der Pol Oscillator \cite{ericaislanantonelo_2022_physicsinformed}, and the classic Four Tank System \cite{ericaislanantonelo_2022_physicsinformed}. While these works demonstrated that it is possible to incorporate physics information into ML dynamics models, they are restricted to Ordinary Differential Equation (ODE) systems which are less general than PDEs are.

In the following sections, we present our approach which incorporates prior knowledge of PDEs into learnt dynamics models for the purpose of manipulating acoustic waves. In our work, we prioritize realizability, with the goal of enabling wave manipulation with real robots.

\section{Preliminaries}
\label{sec:prelim}

In this section, we introduce notations and provide an overview of the problem. The formal problem definition and its solution are provided in Sections \ref{sec:probdef} and \ref{sec:propsol}.

\subsection{Notations}

We denote $t\in \mathbb{R}^+$, $x\in \mathbb{R}^{d_x}$ and $a(t)\in \mathbb{R}^{d_a}$ as the time coordinate, the space coordinate, and the agent's action, respectively. The partial derivative of a function w.r.t time is denoted by $\partial_t$. A differential operator, acting on a function $\f(x, t)$ in a domain $x\in \Omega$, is denoted by $\mathcal{N}(\f; d(t), \ell(x), t)$, where $d(t)$ and $\ell(x)$ represent functions controllable and uncontrollable by the agent, respectively. Both $d(t)$ and $\ell(x)$ can affect the solution to the PDE, $\partial_t \f(x, t)=\mathcal{N}\bigl(\f(x, t); d(t), \ell(x), t\bigr)$. We represent the state of a robot observed at a particular time $t_i$ as $d(t_i)$ whereas $a(t)$ represents actions applied over a time span $t\in\left[t_i, t_{i+1}\right]$.


\subsection{Wave-Robot Coupled Environment}
The simulated environment we consider in this work is defined by a PDE $\mathcal{N}$ for environment dynamics and an ODE $F$ for the robot dynamics. The agent indirectly interacts with the environment through its actions, $a(t)$, which influence the robot, $d(t)$. The latter represents a dynamically changing boundary condition within the environmental domain, which evolves according to a dynamical control system, $F$:
\begin{align}
     &\partial_t \f(x, t) = \mathcal{N}\bigl(\f(x, t);d(t), \ell(x), t\bigr)&\text{PDE}\label{eq:truePDE}\\
     &a(t) = \pi\bigl(\f(x, t), d(t)\bigr)&\text{Policy}\label{eq:policy}\\
     &\dot{d}(t) = F(d(t), a(t))&\text{ODE}\label{eq:design_dynamics}
\end{align}
where $\pi$ is an agent's policy. In this setting the agent controls a low dimensional state of the robot, which in turn affects the function $\f(x, t)$ in the domain $\Omega$. In the sequel we formally define an optimization problem for sample-based derivation of an optimal $\pi$ from a partially observable function $\f(x, t)$. In the interaction model, the ODE controls the PDE creating nested dynamics with two different time scales. 

\subsection{Sparse Robotic Actuation}

The separation between the agent actions, $a(t)$, and the environment $\f$, through the robot, $d(t)$, allows for a physical realization of control of PDEs with a spatially sparse control. 
This is because it is not usually possible to construct a controller which is capable of directly acting on a PDE-governed environment at every point in space. 


In this work, we consider a robotic controller which is capable of manipulating acoustic waves by inducing scattering. Specifically, the robot consists of a number of cylindrical scattering devices that the agent can control through adjustment of their position, radii, or both. In turn, the scattered wave energy produced by reflection of incident waves upon the scatterers can produce complex wave interference patterns, resulting in energy focusing or suppression \cite{amirkulova_2020_the}.

\subsection{Wave Sensing}





In practice, a sensor which is capable of fully observing wave functions is not possible. Instead, real time total displacement and pressure fields can be observed on discretized grids through sensors such as microphone arrays and  acoustic cameras \cite{bocanegra2022_acoustic_camera}. In our simulated environment, the pixels of the sensor images are obtained by observing the displacement of the wave function at points on a uniformly spaced grid over $\Omega$. We assume the agent can only partially observe the wave function, $\f(x, t)$, via its sensors, which we  denote as images: $\mbox{Sensor}\bigl(\f(x, t)\bigr) = X(t)\in \mathbb{R}^{d_1\times d_2}$.






In the next section, we define the control problem, and then propose a tractable method for its solution.


\section{Problem Definition}
\label{sec:probdef}


In the fully observable case, we can formally define the control problem as minimization of a cost function in Eq. \eqref{eq:fullyobservableloss} for $t\in [t_i, t_f]$, subject to the dynamics $\mathcal{N}$ of $\f$ and dynamics $F$ of $d$. A shorthand notation is used $\f = \f(x, t)$:
\begin{align}    
    &\min_{a(\cdot)}\;
    \int_{t_i}^{t_f}||\mathcal{T}(\f(x, t')) - \sigma^*(t')||^2dt' \label{eq:fullyobservableloss}, \\
    &\qquad\mbox{s.t. }
    \begin{cases}
        \partial_t\f = \mathcal{N}(\f; d(t), \ell(x), t),\\
        \dot{d}(t) = F(d(t), a(t)).
    \end{cases}\label{eq:coupledDE}
\end{align}
where $\mathcal{T}(\zeta(x,t))=\sigma(t)$ is a function that computes a signal we wish to predict and control, such as scattered energy, at a particular location or in the entire domain, $\Omega$. The goal of this control problem is to select actions which influence $\f$ to posses properties defined by a reference signal $\sigma^*(t)$.

This problem is challenging, because of the coupling between ODE and PDE in the constraint in Eq. \eqref{eq:coupledDE}, and the complexity of control of an entire function, $\f$. Moreover, wave propagates at a much higher speed in comparison to the responsiveness of a physically realizable robot. That creates two different time scales in the above-mentioned coupling, which requires a particular consideration.   

We assume the robot knows the dynamics, $F$, of its scatterers which define the time-varying boundary conditions for the wave, $\f$. A direct solution to this problem is intractable, so we propose to learn a low-dimensional dynamical control system of $\f$, which captures the essential properties of $\f$, and to utilize this low dimensional dynamics for the design of control law.






\section{Proposed Solution}
\label{sec:propsol}
We propose to solve the problem defined in Section \ref{sec:probdef} in two stages. First, we learn a 1D dynamical control system, $\mathbf{z}(\bar{x}, t)$, which captures the essential properties of full dynamics of wave $\f$, such as energy distribution in space, scattering, excitation, and dissipation. Second, we derive optimal actions to control the energy distribution of $\f$ via its 1D model $\mathbf{z}(\bar{x}, t)$, which is used as a surrogate in the optimization of Eq. \eqref{eq:fullyobservableloss}. A shorthand notation is used $\mathbf{z} = \mathbf{z}(\bar{x}, t)$.

\subsection{Learning the Reduced Dimensional Model}
\begin{figure}[t!]
\centering
\includegraphics[width=\linewidth]{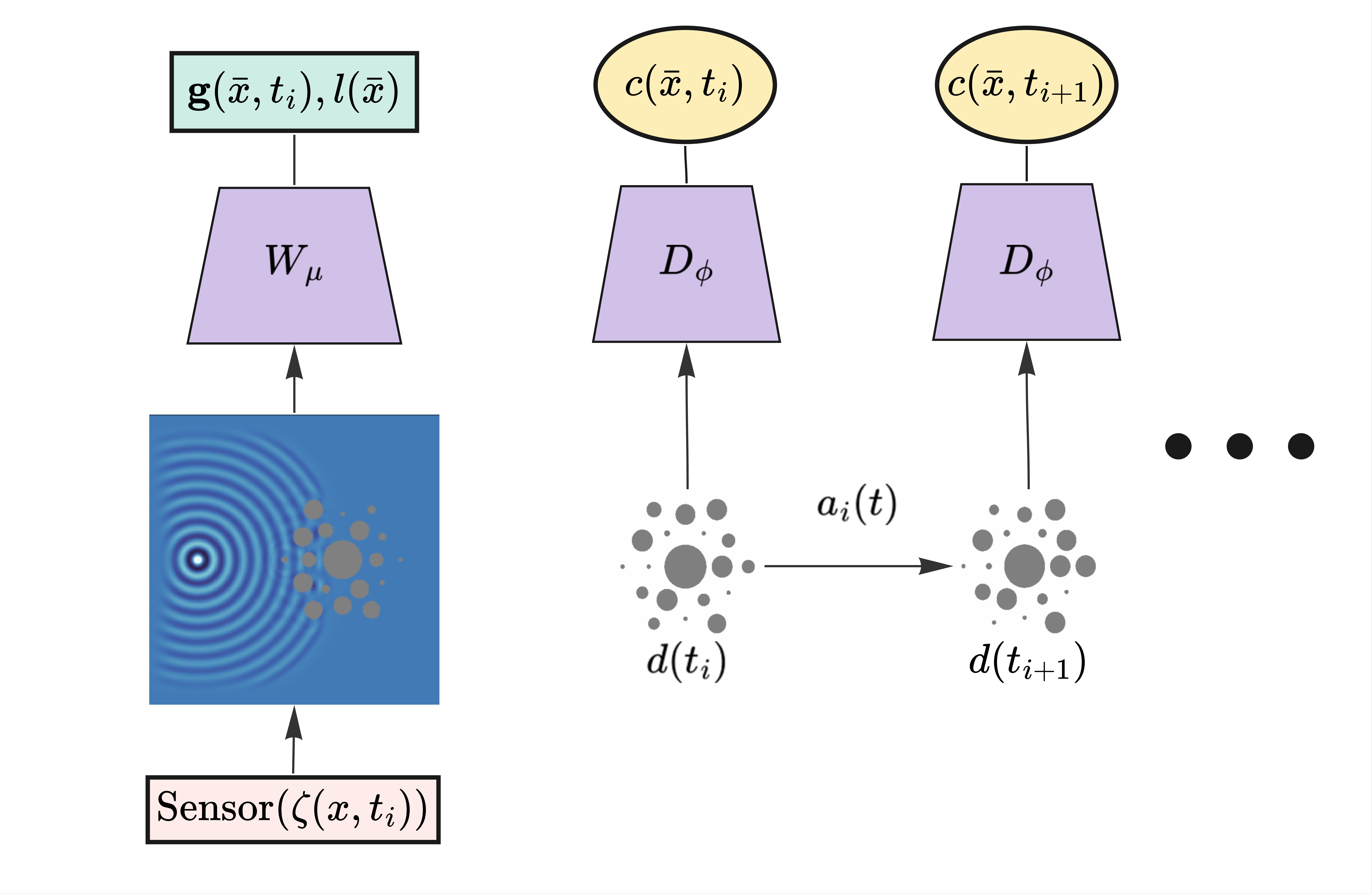}
\caption{Encoding Scheme. Left: Sensor observation $X(t_i)$ of the PDE $\f$ is encoded by $W$ to the latent initial condition $\mathbf{g}(\bar{x}, t_i)$ and exogenous function $l(\bar{x})$. Right: sequences of robot configurations, $d(t_i), d(t_{i+1}), \cdots$, produced by agent's actions, $a_i(t)$ are individually encoded to latent control functions $c(\bar{x}, t_i), c(\bar{x}, t_{i+1}), \cdots$.}\label{fig:Encoders}
\end{figure}

Given an initial sensor observation, $X(t_i)$, robot state $d(t_i)$, and actions $a(t)$, the agent constructs $\mathbf{z}$ for $t \in [t_i, t_f]$, a 1D representation of the true function, $\f$, in the reduced spatial coordinates, $\bar{x}\in\Gamma\subset\mathbb{R}$. Construction of $\mathbf{z}$ is accomplished by way of two encoders shown in Figure \ref{fig:Encoders} which output functions defined over $\Gamma$: the wave encoder, $W$, and the robot encoder, $D$:
\begin{align}
&W_{\mu}:X(t)\rightarrow [\mathbf{g}(\bar{x}, t), l(\bar{x})]&\text{(Wave Encoder)} \label{eq:wave-encoder}\\
&D_{\phi}:d(t)\rightarrow c(\bar{x}, t)&\text{(Robot Encoder)} \label{eq:design-encoder}
\end{align}
where $c(\bar{x}, t)$ and $l(\bar{x})$ represent the environment components which are controllable and uncontrollable by the agent's actions, respectively. This way $c(\bar{x}, t)$ and $l(\bar{x})$ correspond to $d(t)$ and $\ell(x)$, respectively, in the original environment, $\f$. This separation is beneficial because it allows for explicit representation of energy excitation and dissipation properties of the original environment (cf., Figure \ref{fig:main_scheme}) within $l(\bar{x})$. Additionally, $W_{\mu}$ outputs $\mathbf{g}(\bar{x}, t)$, which defines the initial conditions of $\mathbf{z}$. We denote the parameters of $W_{\mu}$ and $D_{\phi}$ by $\theta\doteq[\mu, \phi]$.

Given these definitions, we introduce the following optimization problem for $t\in [t_i, t_f]$: 
\begin{align}
    &\min_{\theta}\Biggl\{ \int_{t_i}^{t_f}||\mathcal{T}\bigl(\mathbf{z}(\bar{x}, t')\bigr) - \mathcal{T}(\zeta(x, t')||^2dt'\label{eq:ControlCostLatent}\\
    &+ \int_{t_i}^{t_f}\!\!\!\!\int_\Gamma ||\partial_{t} \mathbf{z}(\bar{x}, t') - \mathcal{N}^\Gamma(\mathbf{z}(\bar{x}, t'); c_{\phi}(\bar{x}, t'), l_{\mu}(\bar{x}), t')||^2 d\bar{x}dt' \label{eq:PhysicsLossLatent}\\
    &\qquad\qquad\;+ \int_\Gamma ||\mathbf{z}(\bar{x}, t_i) - \mathbf{g}_{\mu}(\bar{x}, t_i)||^2 d\bar{x}\label{eq:ICLatent}\Biggr\}\\
    &\qquad\mbox{s.t. }\;\dot{d}(t) = F(d(t), a(t))\nonumber,
\end{align}
where $\mathcal{N}^{\Gamma}$ is a PDE similar to $\mathcal{N}$ that operates on a 1D space $\Gamma$ instead of $\Omega$. Eq. \eqref{eq:ControlCostLatent} is the prediction loss for scattered energy. Eq. \eqref{eq:PhysicsLossLatent} and Eq. \eqref{eq:ICLatent} are the PDE consistency and the initial condition losses, respectively. Notably, $\mathbf{z}$ implicitly depends on $\theta$ through learnable $\mathbf{g}_{\mu}$, and $c_{\phi}$ and $l_{\mu}$.

We chose a deep convolutional neural network to represent $W_\mu$, in Figure \ref{fig:Encoders} (left), which maps the pixels of the sensor observations $X(t)$ onto functions over $\Gamma$. Similarly, the robot encoder is implemented by an artificial neural network $D_\phi$, in Figure \ref{fig:Encoders} (right), maps the robot's trajectory to latent control functions. Since the dynamics of the robot in Eq. \eqref{eq:design_dynamics} are known ahead of time, $D_\phi$ can construct $c(\bar{x}, t)$ from an initial robot state $d(t_i)$ and $a(t)$.

In our method, $\mathbf{z}$ is constructed using numerical integration of $\mathcal{N}^{\Gamma}$. Assuming prior knowledge of the governing equation of $\f$ in lower dimensional space, the encoded initial conditions, boundary conditions, and exogenous forces specify a solution $\mathbf{z}$. Relying on numerical integration to generate $\mathbf{z}$ has the unique benefit of guaranteeing that it is a solution to $\mathcal{N}^{\Gamma}$, while introducing no trainable parameters aside from the encoders $W_\mu$ and $D_\phi$. This is in contrast to the approaches that require an additional network (e.g., NODE) to represent the dynamics $\mathcal{N}^{\Gamma}$ of $\mathbf{z}$, which increases sample and computational complexity.

Our method requires the following data for training:
\begin{align}
\mathcal{I}=\{\bigl(X(t_i), d(t_i), a(t), \sigma(t)\bigr)\}_{k=1}^N,\label{eq:dataset}   
\end{align}
where $X(t_i)$, $d(t_i)$, $a(t)$, and $\sigma(t)$, are a sensor observation and a robot state observed at time $t_i$, and actions and corresponding target signal in time period $t\in \left[t_i, t_{f} \right]$. Algorithm~\ref{alg:cPILS_NI} summarizes the learning of physics-informed model. 

\begin{algorithm}[h!]
\caption{Learning Physics-Informed Model}
\begin{algorithmic}[1]\label{alg:cPILS_NI}

\REQUIRE Training data, $\mathcal{I}$, (cf., Eq. \eqref{eq:dataset}), parameters of $W_{\mu}$ and $D_{\phi}$, denoted by $\theta = (\mu, \phi)$, and design dynamics $F$.

\STATE Randomly initialize $\theta$
\REPEAT
    \setlength\itemsep{0.5em}
    \STATE \textbf{Sample} a data point from $\mathcal{I}$:\hfill\COMMENT{Eq. \eqref{eq:dataset}}
    $X(t_i), d(t_i), a(t), \sigma(t)\sim \mathcal{I}$
    \STATE \textbf{Encode} a Sensor Reading (wave image):\hfill\COMMENT{Eq. \eqref{eq:wave-encoder}}
    $\mathbf{g}_{\mu}(\bar{x}, t_i), l_{\mu}(\bar{x}) = W_{\mu}(X(t_i))$
    \STATE \textbf{Integrate} a Design Trajectory from $d(t_i)$:\\
    $d(t) = d(t_i) + \int_{t_i}^{t_f}F(d(t'), a(t'))dt'$
    \STATE \textbf{Encode} a Design Trajectory:\hfill\COMMENT{Eq. \eqref{eq:design-encoder}}
    $c_\phi(\bar{x}, t) = D_\phi(d(t))$
    \STATE \textbf{Integrate} a Latent Trajectory from $\mathbf{g}(\bar{x}, t_i)$:\vspace{-0.5cm}\\
    \begin{align*}
        &\mathbf{z}_{\theta}(\bar{x}, t) = \\
        &\quad\mathbf{g}_{\mu}(\bar{x}, t_i)+ \int_{t_i}^{t} \mathcal{N}^{\Gamma}(\mathbf{z}(\bar{x}, t'); c_{\phi}(\bar{x}, t'), l_{\mu}(\bar{x}), t')dt'
    \end{align*}
    \vspace{-0.5cm}
    \STATE \textbf{Calculate} the Prediction Loss:\hfill\COMMENT{Eq. \eqref{eq:ControlCostLatent}}\\
    $\mbox{loss}(\theta) = \int_{t_i}^{t_f} \bigl(\mathcal{T}\bigl(\mathbf{z}_{\theta}(\bar{x}, t')\bigr) - \sigma(t') \bigr)^2dt'$
    \STATE \textbf{Perform Gradient Descent} on  $\mbox{loss}(\theta)$
\UNTIL{Convergence}
\RETURN $W^*_{\mu}$, $D^*_{\phi}$
\end{algorithmic}
\end{algorithm}
Once the robot learns how its sparse control affect the wave, $\f$, through $\mathbf{z}$, optimal actions can be selected according to Model Predictive Control, as explained below.

\subsection{Model Predictive Control of Wave Energy}

We apply MPC \cite{bakaravc2018_random_shooting,RICHALET1978_MPC, MORARI1988_MPC} for control of wave energy over the time interval $[t_i, t_f]$, by dividing a control trajectory, $a(t)$, to a sequence of $N_a$ piece-wise constant actions, $a(t)=\{a_{\tau} : \mbox{ if }t_\tau\le t < t_{\tau+1}\,\forall \tau \}$. This time discretization leads to a series of time intervals $\Delta_{\tau} = (t_{\tau+1}-t_{\tau})$, with boundary conditions $t_{\tau = 1} = t_i$ and $t_{\tau = N_a + 1} = t_f$, which reflects the slow time scale of robotic actuation in comparison to the high speed of wave propagation. The objective below with the reference control signal $\sigma^*(t)$, allows for the derivation of the optimal control sequence $\{a^*_{\tau}\}_{\tau=1}^{N_a}$:   
\begin{align}
 &\underset{\{a_{\tau}\}_{\tau=1}^{N_a}}{\min} \int_{t_i}^{t_f}||\mathcal{T}(\mathbf{z}(\bar{x}, t')) - \sigma^*(t')||^2dt'
    + \beta\sum_{\tau=1}^{N_a}||a_{\tau}||^2\Delta_{\tau}    \label{eq:latent_control_cost}\\
    &\mbox{s.t.} \begin{cases}
        \partial_{t}\mathbf{z}(\bar{x}, t) = \mathcal{N}^\Gamma(\mathbf{z}(\bar{x}, t); c_{\phi^*}(\bar{x}, t), l_{\mu^*}(\bar{x}), t)\\
        \mathbf{z}(\bar{x}, t_i) = \mathbf{g}_{\mu^*}(\bar{x}, t_i)
    \end{cases}
\end{align}
where the constraints are the latent 1D dynamics with the optimal parameters $\theta^* = [\mu^*, \phi^*$] in Eq.~\eqref{eq:PhysicsLossLatent}-\eqref{eq:ICLatent}, in addition to F, which is known to the agent. The agent's power is constrained through hyperparameter $\beta$.


To the best of our knowledge, this proposed solution has not been studied prior to our work, while the most relevant works \cite{ericaislanantonelo_2022_physicsinformed, nicodemus_2022_physicsinformed} considered ODEs only, which is a restricted class of dynamics. Moreover, there are no efficient data-driven methods for controlling acoustic PDEs in open space with sparse control and partially observable state.

\section{Experiments}
\label{sec:experiments}

The goal of our experiments is to validate that the AEM approach can effectively exploit prior knowledge about the physical properties of a system governed by an acoustic PDE. We demonstrate this by first showing that the model can predict scattered energy in the domain for action sequences extending far beyond its training horizon. Next, we show that AEM can be effectively used with MPC on two benchmark tasks in acoustics: energy suppression and focusing. We compare our method to a ML approach (NODE) and a ground truth solution derived by GBO. 

\subsection{Setting}

The robotic actuator we consider in this environment is a configuration of cylindrical scatterers. We assume that an agent has the ability to manipulate by $a(t)$, the radii and/or position of $d(t)$, which create time-variant boundary conditions for the domain, $\Omega(d(t))$ of wave propagation:
\begin{align*}
    &\partial_t^2\zeta(x, t) = c^2\partial_x^2\zeta(x, t) + f(x, t) \in \Omega(d(t))\\
    &\dot{d}(t) = F(d(t), a(t)),
\end{align*}
where $\f(x, t)$ and $d(t)$ evolve in two different timescales: fast  and slow for wave and robot, respectively.

We conduct the experiments in a simulated open space with size $x \in \left[\SI{-15.0}{\meter}, \SI{15.0}{\meter} \right] \times \left[\SI{-15.0}{\meter}, \SI{15.0}{\meter} \right]$, where 
the excitation source creates a spherical wave in the domain (cf., Figure \ref{fig:main_scheme}). Simulation of open space without wave reflection from the boundaries is achieved through a 
PML \cite{berenger_1994_a}, which dissipates wave energy as it leaves $\Omega$. Within $\mathcal{N}^\Gamma$ we include a trainable PML through $l(\bar{x})$ which allows our agent to explicitly model energy dissipation.

\subsection{Robot Actuation}
We vary the number of scatterers, denoted as $M$, and test several configurations for robot actions. We use \textbf{R}, \textbf{P}, and \textbf{F} to represent "Radii Adjustment", "Positional Adjustment", and "Full Adjustment" respectively:
\begin{enumerate}
    \item \textbf{Radii Adjustment (R)} (Ring Configuration): This setup features $M = 19$ scatterers arranged in a ring formation with fixed positions. The actions in this experiment involve solely adjusting the radii of the scatterers.
    \item \textbf{Positional Adjustment (P)}: Here, scatterers have fixed radii, and the agent can only change their positions. We tested this configuration with $M = 1$, $2$, and $4$ scatterers.
    \item \textbf{Full Adjustment (F)}: In this scenario, the agent can modify both the positions and radii of the scatterers. We demonstrated performance for $M = 2$ in this configuration.
\end{enumerate}



\subsection{Evaluation Tasks}
We evaluate our model in both prediction and control.
\begin{itemize}
    \item \textbf{Long-Term Prediction}: We compare predicted scattered energy over a horizon of 200 steps (much beyond the training horizon) between our model, the baseline model, and the ground truth. 
    \item \textbf{Focusing}: Concentrate the scattered energy within the upper-right quadrant of the domain by the robots actions. 
    \item \textbf{Suppression}: Minimize the total scattered energy within the domain by the robots actions.
\end{itemize}
These tasks represent standard benchmarks and are frequently used in acoustics research \cite{amirkulova_2020_the, norris_2008_acoustic}. We use Model Predictive Control in both manipulation tasks.

\begin{table}
\centering
\vspace{0.2cm}
\caption{Steady-state scattered energy in the simulated environment. The steady-state energy is measured from $0.10$ to $0.20$ seconds of time in the environment. Statistics are computed over 12 runs with randomized initial robot conditions and fixed source location at $(-10.0, 0.0)$.}
\label{tab:results_summary}
\begin{tabularx}{\linewidth}{l cccc}
\toprule
\textbf{Method} & \textbf{Random} & \textbf{NODE} & \textbf{AEM} & \textbf{GBO} \\
\midrule
\multicolumn{5}{c}{\textbf{Focusing (Higher is Better)}} \\
\textbf{M=1 (P)}     & $0.12 \pm 0.03$ & $0.17 \pm 0.07$ & $\mathbf{2.47 \pm 1.91}$ & N/A \\
\textbf{M=2 (P)}     & $0.35 \pm 0.21$ & $1.22 \pm 1.22$ & $\mathbf{7.54 \pm 2.24}$ & N/A \\
\textbf{M=4 (P)}     & $1.05 \pm 0.46$ & $1.45 \pm 0.92$ & $\mathbf{5.94 \pm 1.90}$ & N/A \\
\textbf{M=2 (F)}     & $0.87 \pm 0.60$ & $6.19 \pm 1.51$ & $\mathbf{8.49 \pm 1.37}$ & N/A \\
\midrule
\multicolumn{5}{c}{\textbf{Suppression (Lower is Better)}} \\
\textbf{M=1 (P)}     & $3.00 \pm 0.72$ & $2.86 \pm 0.53$ & $1.08 \pm 0.18$ & $\mathbf{1.01}$ \\
\textbf{M=2 (P)}     & $6.87 \pm 1.89$ & $4.40 \pm 0.84$ & $\mathbf{1.98 \pm 0.38}$ & $2.13$ \\
\textbf{M=4 (P)}     & $10.04 \pm 1.84$ & $10.16 \pm 2.11$ & $3.78 \pm 0.74$ & $\mathbf{3.53}$ \\
\textbf{M=2 (F)}     & $5.41 \pm 1.33$ & $3.48 \pm 1.01$ & $\mathbf{0.38 \pm 0.02}$ & $0.38$ \\
\bottomrule
\end{tabularx}
\end{table}

\subsection{Results}

\subsubsection{Long-Term Prediction of Scattered Energy}
We assess the robustness of our model by comparing its long-term prediction capabilities to a NODE-based approach, \begin{figure}
\includegraphics[width=\linewidth]{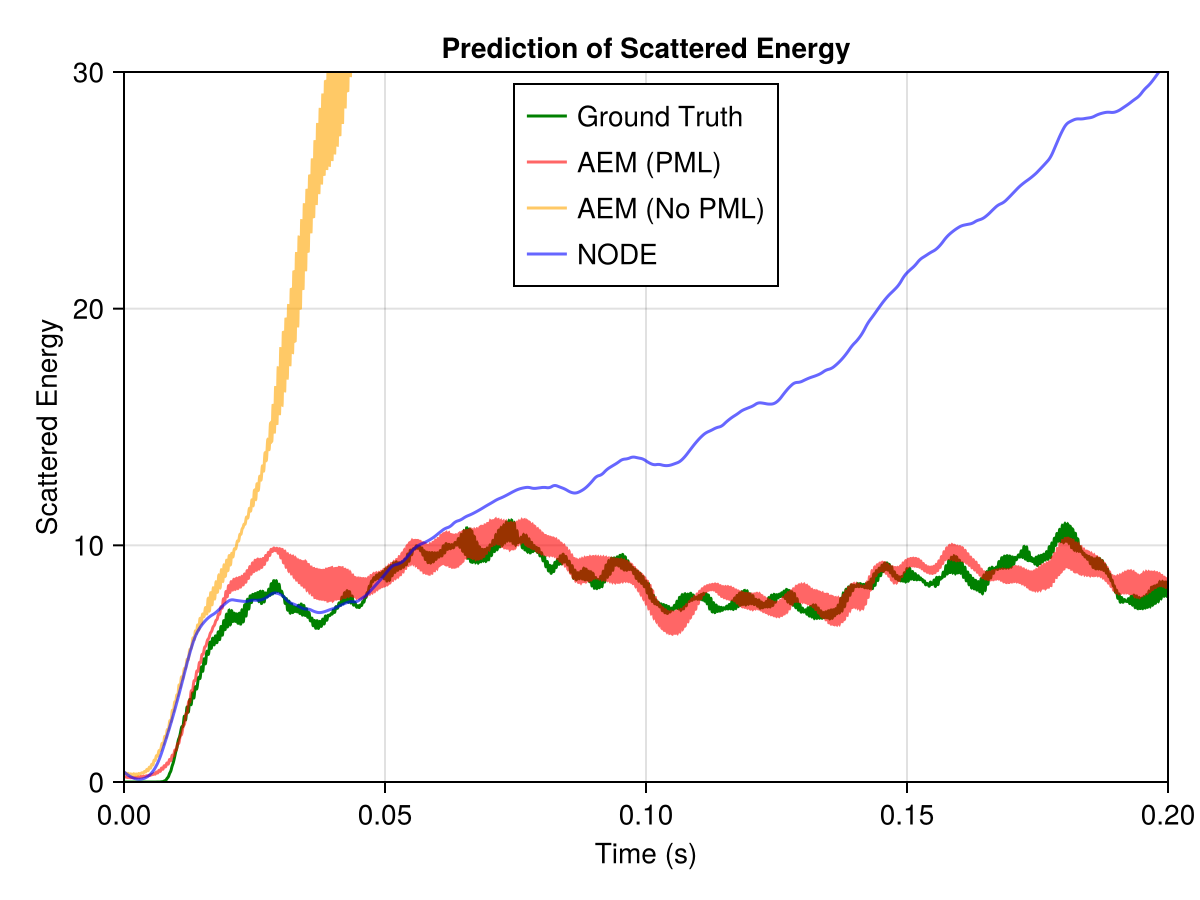}
\caption{Long-term prediction of scattered energy over an episode of 0.2 seconds (200 action steps) for the ring configuration (\textbf{R}). The training horizon for the models is 20 actions, which demonstrate generalization of the model to unseen in training horizons.}\label{prediction_plot}
\end{figure} as well as an AEM-based approach that does not incorporate a trainable PML inside $\mathcal{N}^\Gamma$. Figure \ref{prediction_plot} shows that AEM (PML) follows the ground truth scattered energy compared to NODE and AEM (No PML) which diverge. Our model without a PML experiences rapid energy buildup because it is unable to dissipate energy in $\mathbf{z}$. The NODE model also diverges from the ground truth signal and provides no pathway to explicitly model energy dissipation. These results highlight the critical role of including a PML in $\mathcal{N}^\Gamma$ when $\f$ occurs in open space. Table \ref{tab:results_summary} shows that AEM achieves similar results to the ground truth results by GBO in the 'Suppression' task, when a semi-analytical solution by GBO is known in the literature, otherwise it is marked 'N/A'.\\

\subsubsection{Focusing and Suppression of Scattered Energy}
\begin{figure}
\includegraphics[width=\linewidth]{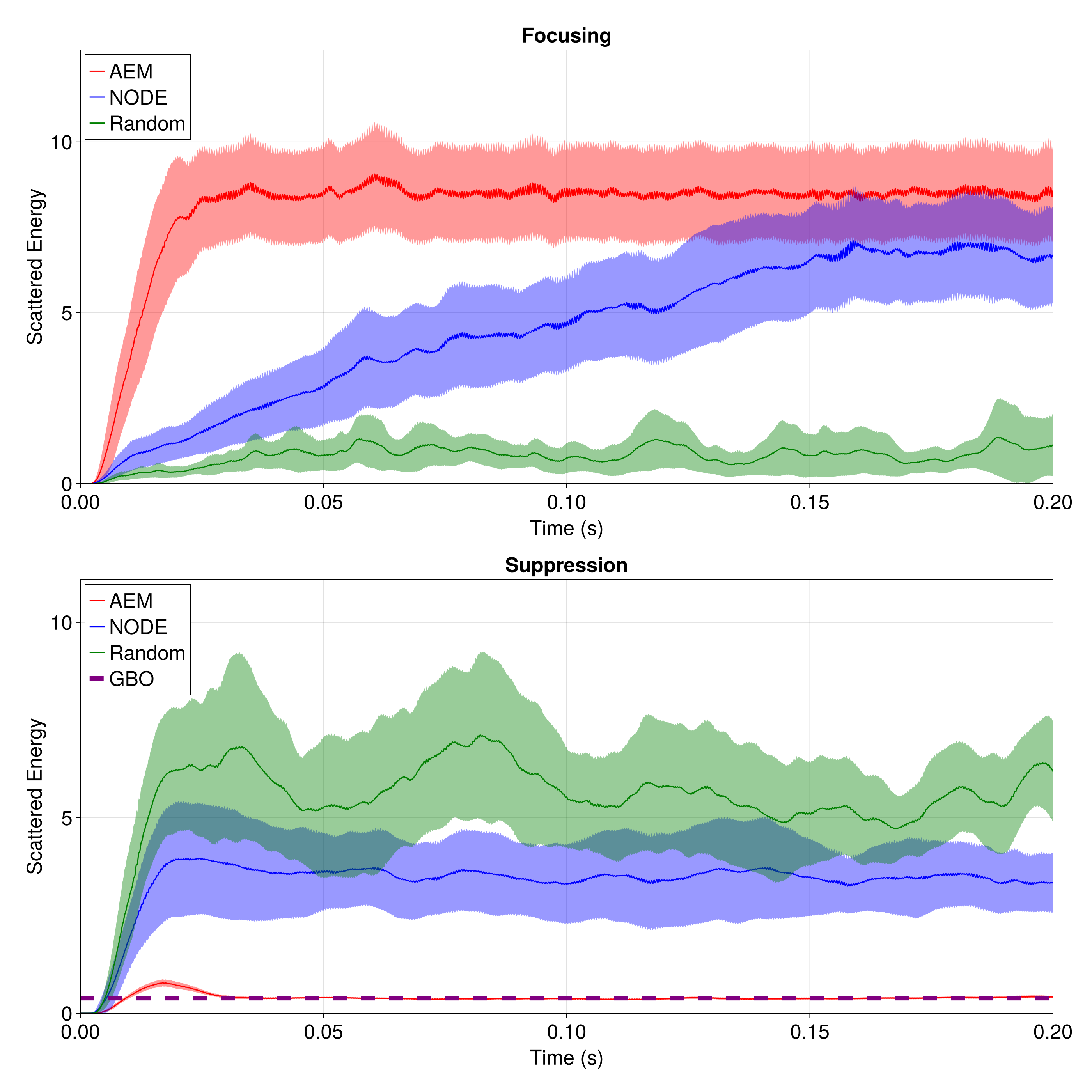}
\caption{Focused and Suppressed Scattered Energy.  Both the positions and radii of two fully adjustable scatterers are varied (F, M=2). Source is fixed at location (-10, 0) throughout the simulations. Shading represents ± 0.5 standard deviation from the mean. The mean (solid lines) and variance are calculated with 12 runs with randomized initial conditions. The dashed line is for the ground truth by GBO.}\label{fullyAdj_mean}
\end{figure}

We utilize MPC through optimization of Eq. \eqref{eq:latent_control_cost} for both benchmark tasks. Figure \ref{fullyAdj_mean} shows the results for the fully adjustable configuration with two scatterers (F, M=2). The plot demonstrates the advantages of our approach in both focusing (top) and suppressing (bottom) scattered energy. In terms of steady-state performance, our method achieves higher focused energy and lower suppressed energy. Furthermore, our approach exhibits better transient response time, reaching steady-state faster than the other approaches. Similar trends were observed across other scatterer configurations.

We compare the performance of our model against the NODE-based approach with both models having an equal parameter count of 2.7 million. Table \ref{tab:results_summary} summarizes the results of our MPC experiments. A random policy, and gradient-based optimization approaches are also included as part of the comparison. Our AEM model achieves good performance on both benchmarks: for focusing, it yields higher focused energy while maintaining comparable error rates; for suppression, it attains lower suppressed energy than NODE and Random control with the added benefit of reduced error compared to other methods. For some of the cases AEM achieves slightly better performance than the steady-state GBO solution.




\section{Conclusions}
\label{sec:conclusions}

This work extends classical object manipulation in robotics to the domain of acoustic waves.
It opens new 
research opportunities at the intersection of robotics and acoustics, with potential applications in such areas as ultrasound cutting, energy harvesting, and the design of new acoustic materials. This work provides proof of concept for the implementation of a physically realizable robot for wave manipulation. Importantly, the methodology is not limited to acoustics—it can be adapted to other domains governed by PDEs.

We highlight the unique features of our approach, including sparse robotic actuation under partial observability and a fully interpretable, physics-constrained solution. This allows for a trainable PML embedded within the AEM’s latent space ensuring robust and accurate long-term predictions. The proposed AEM allows to find a solution, which can be derived only in particular cases with GBO. That makes AEM a strong candidate in robot learning for PDE manipulation, where interpretable and reliable methods are required.






\section*{Acknowledgements}

The authors thank P.~E.~Leser from NASA for the valuable suggestions. 
ST was supported in part by NSF (2246221) and PAZI (195-2020).
TS was supported in part by the Koh Family Scholarship.
TS and NS acknowledge the RSCA Awards.
Part of this work was done when TS and NS were graduate students at SJSU.

\bibliographystyle{IEEEtran}
\bibliography{bibliography}



\end{document}